\documentclass[12pt,a4paper]{article}
\usepackage{amsmath}
\usepackage{amsfonts}
\usepackage{amssymb}
\usepackage{graphicx}
\usepackage{verbatim}
\usepackage{amsmath,amsfonts,enumerate}
\usepackage{framed,multirow}
\usepackage{algorithm}
\usepackage{algpseudocode}
\usepackage{amssymb}
\usepackage{latexsym}
\usepackage{subcaption}
\usepackage{booktabs}
\usepackage{url}
\usepackage{xcolor}
\usepackage{color,soul}
\usepackage[colorlinks,linkcolor=red,anchorcolor=blue,citecolor=green]{hyperref}
\numberwithin{equation}{section}
\providecommand{\keywords}[1]
{
  \small	
  \textbf{\textit{Keywords---}} #1
}
\title{Conservative Physics-Informed Neural Networks for Non-Conservative Hyperbolic Conservation Laws Near Critical States}

\author{Reyna Quita$^{a,1}$, Yu-Shuo Chen$^{a,2}$, Hsin-Yi Lee$^{b}$, Alex C. Hu$^{a,3}$, John M. Hong$^{a,4}$   \\
        \small $^{a}$Department of Mathematics, National Central University, Jhongli City, Taoyuan, 32001,Taiwan\\
        \small $^{b}$Department of Mathematics, National Cheng Kung University, Tainan 70101, Taiwan \\
        \small $^{1}$reynaquita2905@gmail.com, $^{2}$formosa1502@gmail.com, $^{b}$apostol2000@hotmail.com\\
        \small $^{4}$calexhu@gmail.com, $^{5}$jhong@math.ncu.edu.tw \\
}
\date{} 

\begin{document}
\maketitle

\begin{abstract}
In this paper, a modified version of conservative Physics-informed Neural Networks (cPINN for short) is provided to construct the weak solutions of Riemann problem for the hyperbolic scalar conservation laws in non-conservative form. To demonstrate the results, we use the model of generalized Buckley-Leverett equation (GBL equation for short) with discontinuous porosity in porous media. By inventing a new unknown, the GBL equation is transformed into a two-by-two resonant hyperbolic conservation laws in conservative form. The modified method of cPINN is invented to overcome the difficulties due to the discontinuity of the porosity and the appearance of the critical states (near vacuum) in the Riemann data. We experiment with our idea by using a deep learning algorithm to solve the GBL equation in both conservative and non-conservative forms, as well as the cases of critical and non-critical states. This method provides a combination of two different neural networks and corresponding loss functions, one is for the two-by-two resonant hyperbolic system, and the other is for the scalar conservation law with a discontinuous perturbation term in the non-convex flux. The technique of re-scaling to the unknowns is adopted to avoid the oscillation of the Riemann solutions in the cases of critical Riemann data. The solutions constructed by the modified cPINN match the exact solutions constructed by the theoretical analysis for hyperbolic conservation laws. In addition, the solutions are identical in both conservative and non-conservative cases. Finally, we compare the performance of the modified cPINN with numerical method called WENO5. Whereas WENO5 struggles with the highly oscillation of approximate solutions for the Riemann problems of GBL equation in non-conservative form, cPINN works admirably.
\end{abstract} \hspace{10pt}

\section{Introduction}
In this paper, we provide a modified version of conservative physics-informed neural networks (cPINN for short) to construct the weak solutions of hyperbolic conservation laws in non-conservative form. We use the generalized Buckley-Leverett (GBL for short ) equation with variable porosity in porous media to demonstrate our results. The GBL equation is read as
\begin{equation}\label{eqn:ori_GBL}
    \phi\tilde{u}_t + \tilde{f}(\tilde{u},x)_x=0,\quad (x,t)\in\mathbb{R}\times\mathbb{R}^+,
\end{equation}
where $\tilde{u}=\tilde{u}(x,t)$ is the saturation of the water, $\phi=\phi(x)>0$ is the porosity of the medium, the positive constant $M$ is the water over oil viscosity ratio, and the flux $\tilde{f}(\tilde{u},x)$ is defined as 
\begin{equation}\label{eqn:tilde_flux}
    \tilde{f}(\tilde{u},x) = \frac{\tilde{u}^2}{\tilde{u}^2+M(1-\tilde{u})^2},\quad 0\leq \tilde{u}\leq 1,\quad M>0. 
\end{equation}
The equation (\ref{eqn:ori_GBL}) models the motion of water flow in oil-water flows in a porous medium. The equation \eqref{eqn:ori_GBL} is a scalar hyperbolic equation in non-conservative form.
It can be re-written in a conservative form by the change of variable, as we let
\begin{equation}\label{eqn:changing}
    u(x,t):=\phi(x)\tilde{u}(x,t).
\end{equation}
Then we obtain the following equation in conservative form
\begin{equation}\label{eqn:GBL}
u_t +f(u,\phi(x))_x = 0,\quad (x,t)\in\mathbb{R}\times\mathbb{R}^+,\\ 
\end{equation}
where the flux $f(u,\phi(x))$ is defined as 
\begin{equation}\label{eqn:flux}
    f(u,\phi(x)) = \frac{u^2}{u^2+M[\phi(x)-u]^2},\quad 0\leq u\leq \phi(x). 
\end{equation}
For the derivation of the Buckley-Leverett equation, we refer the reader to \cite{Duijn2007,Wang2013}. In this paper, we focus on the case that the porosity $\phi$ is a piece-wise constant function so that $\phi\tilde{u}_t$ is not defined in the sense of distribution in PDEs theory when $u$ consists of discontinuities (or shock waves). We provide the application of cPINN to dedicate that the profiles of weak solutions to the Riemann problems of (\ref{eqn:ori_GBL}) and (\ref{eqn:GBL}) are identical, which means that there is no difference for the machines whether the hyperbolic systems are in conservative form or not.

Deep learning (DL for short) has clearly been integrated into many aspects of our daily lives. Among these are self-driving cars, face recognition technology, machine translation, and so forth. In addition, DL is employed in a fairly novel way to solve PDEs while adhering to any physics laws specified by the governed equations as the background knowledge. This concept is first materialized by \cite{pinn} and is known as Physics-informed Neural Networks (PINN for short). As demonstrated in \cite{pinn1,pinn2,pinn3,pinn4,pinn5}, PINN have been implemented auspiciously to solve a wide range of forward and inverse problems of PDEs. Nonetheless, the accuracy of the solution generated by machine is bounded below and the high training costs are some disadvantages of PINN \cite{cpinn}. Other than those two shortcomings, PINN's fundamental limitation is its inability to provide a satisfactory approximation to the PDEs with discontinuous solutions (for example, shock waves) \cite{limit_pinn}.  

\begin{figure}[h]
        \centering
        \includegraphics[scale=0.5]{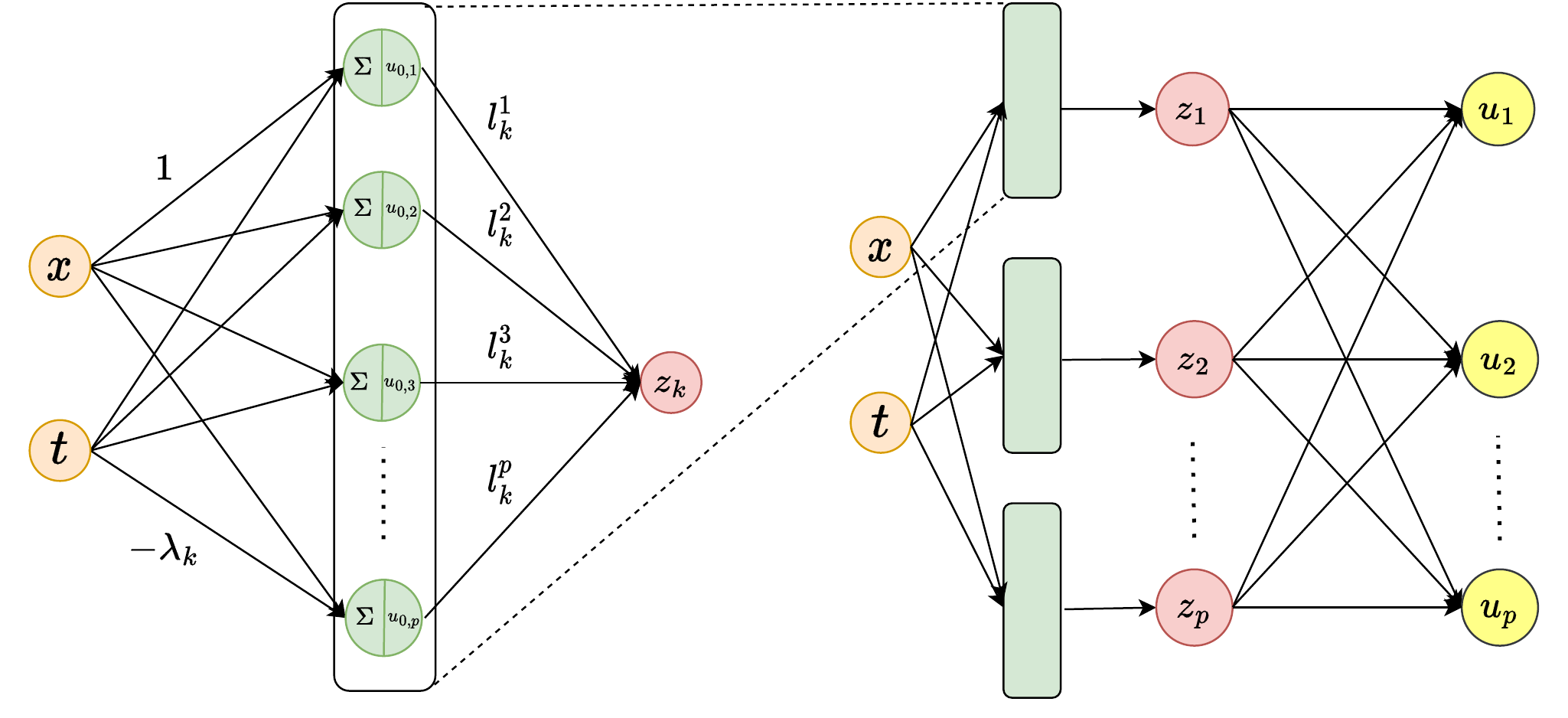}
        \caption{Neural network representation of formula (\ref{eqn:linSOL}).}
        \label{fig:linSOL}
    \end{figure}

The mathematical concept of using the neural networks to construct the solutions of hyperbolic systems of conservation laws is as follows. Consider a linear hyperbolic system with constant coefficients which is written as follows:
\begin{equation}\label{sys:linear}
    \left\lbrace \begin{array}{l}
    u_t+ Au_x=0,\quad (x,t)\in\mathbb{R}\times\mathbb{R}^+,\\
    u(x,0) = u_0(x),\quad x\in\mathbb{R},
    \end{array}\right.
\end{equation}
where $u(x,t)$ is a vector-valued function in $\mathbb{R}^p$ and $A$ is a constant matrix of size $p\times p$.
Suppose that $A$ has $p$ distinct eigenvalues : $\lambda_1<\lambda_2<\cdots<\lambda_p$.
Then the solution of the Cauchy problem (\ref{sys:linear}) is given by
\begin{equation}\label{eqn:linSOL}
    u(x,t) = \sum_{k=1}^{p}\textbf{l}_k^Tu_0(x-\lambda_k t)\textbf{r}_k,
\end{equation}
where $\textbf{l}_k$ and $\textbf{r}_k$ are left and right eigenvectors of $A$ for $1\leq k \leq p$ respectively.
Setting $z_k = \textbf{l}_k^Tu_0(x-\lambda_k t)$. 
Then the formula (\ref{eqn:linSOL}) can be re-written as below
\begin{equation}\label{eqn:linNSOL}
    u(x,t) = \sum_{k=1}^{p}\textbf{l}_k^Tu_0(x-\lambda_k t)\textbf{r}_k=\sum_{k=1}^{p}z_k\textbf{r}_k.
\end{equation}
Thus $u(x,t)$ can be represented as a neural network with initial data being its activated function in Figure \ref{fig:linSOL}.
In general, the activation functions in a neural network are hyperbolic tangent or sigmoid functions. 
With the results of the universal approximation theories in \cite{1989_Cybenko,HORNIK1991251,pinkus_1999},
the initial condition can be approximated by a neural network.
Based on this fact, we believe that PINN can give an approximation solution to a Riemann problem of a nonlinear hyperbolic system. 
As we mentioned that the solution may be constructed by solving two stages of system (\ref{sys:GBL}), 
thus it is worth considering using different architectures of neural networks in each stage.
More precisely, we divide the domain into two sub-domains,
and we use one PINN to solve the problem in one subdomain. 
This strategy can be realized by applying conservative PINN.    

Conservative PINN (cPINN) is an extension of PINN where the algorithm's primary objective is to solve the conservation law. In fact, cPINN is an attempt to address the first two issues of PINN that were previously mentioned—the solution's accuracy and the high training cost. In cPINN, domains are divided into numerous non-overlapping subdomains. In each sub-domain, we consider different neural network architectures, such as ones with a different number of outputs (thus, scalar or system case), various numbers of hidden layers and neurons, distinct sets of hyper-parameters and parameters, different activation functions, different optimization techniques, distinct number of training and interior sample points, and so forth. This will increase our ability to select the best network for each sub-domain. In order to maintain the continuity, the solution in each sub-domain is eventually pieced back together using the proper interface conditions. Despite the fact that the performance of cPINN on hyperbolic conservation laws was not thoroughly investigated in the original paper \cite{cpinn}, we persist to use cPINN on our governed equations considering we have developed a completely distinct purpose for cPINN that allows us to implement various system and scalar architectures in each subdomain. 

Our goal is to  simulate the Riemann problem of equation (\ref{eqn:GBL}) by using a deep machine for the following Riemann problem
\begin{equation}\label{eqn:R_GBL}
    \left\lbrace 
    \begin{array}{l}
    u_t + f(u,\phi(x))_x = 0,\\
    u(x,0) = \left \lbrace \begin{array}{l}
            u_L,\quad x<0,\\
            u_R,\quad x<0,
    \end{array}\right.
    \end{array}\right.
\end{equation}
where $u_L$, $u_R$ are two constant states, and $\phi(x)=\phi_L$ when $x<0$, and $\phi(x)=\phi_L$ when $x>0$.
Following the analysis in \cite{Hong2012},  we augment the equation (\ref{eqn:GBL}) with $\phi_t = 0$. 
Then we obtain the following equivalent Riemann problem of $2\times2$ system of conservation laws
\begin{equation}\label{sys:GBL}
    \left\lbrace 
    \begin{array}{l}
    U_t + F(U)_x = 0,\\
    U(x,0) = \left \lbrace \begin{array}{l}
            U_L,\quad x<0,\\
            U_R,\quad x<0.
    \end{array}\right.
    \end{array}\right.
\end{equation}
where $U:=(u,\phi)^T, F(U):=(f(u,\phi),0)^T$, $U_L=(u_L,\phi_L)^T$,$U_R=(u_R,\phi_R)^T$.

The solution of (\ref{sys:GBL}) consists of the elementary waves form each characteristic wave field, which is standing wave discontinuity from the linear degenerate field and the rarefaction, shock waves from the nonlinear genuinely field. Moreover, there is a path from $U_L$ to $U_R$ consisting of a sequence of corresponding wave curves in $u-\phi$ plane. Based on the results in \cite{Hong2012}, these waves are obtained by solving a two-by-two hyperbolic resonant system followed by a scalar hyperbolic equations with non-convex flux $f(u,\phi_R)$. When solving the Riemann problem (\ref{sys:GBL}), there is a time-independent wave in the solution, which is called the standing wave discontinuity, due to the fact that the first characteristic field is linear degenerate. For the second characteristic field which is genuinely nonlinear, the rarefaction curves are horizontal line in $u-\phi$ plane, thus $\phi$ stays constant on the rarefaction wave curves. Since the solution of (\ref{sys:RieGBL}) consists of a standing wave coming from the two-by-two system, and a time dependent nonlinear wave coming from the scalar equation, according to the construction of the solution, we propose to use two different neural networks, one for the case of system, and the other for the scalar equation respectively. These two neural networks are separated by an interface.
From the observation of the standing shock and the Riemann data of $\phi$, we are able to specify the location of interface, where is on the right hand side and close to the wall $x=0$ (or $t$ axis).

In this paper, we also consider the case that the initial data is critical. 
It means that the initial data is extremely close to zero. In this case, the weak solutions constructed by the original cPINN either have the wrong profile or the speeds are incorrect. To overcome the problem, we propose to re-scale the unknowns $u$ and $\phi$ so that the new flux under the re-scaling becomes non-singular. As we observe, an entropy condition is required we deal with the discontinuous solutions in PINN and cPINN. The well-known entropy conditions that are frequently used in PINN are the ones invented by Oleinik \cite{oleinik1, oleinik2}, Kruzkhov \cite{wpinn}, and the concept of entropy-entropy flux pair \cite{entropy_flux_pair, entropy_flux_pair1} in PDEs. In our framework, Oleinik's entropy condition is used and is needed to be modified to obtain the correct speeds of weak solutions in the critical case. In addition, the choice of the scaling parameters becomes an important issue. Under a suitable choice of re-scaling parameters and the re-scaling technique, we are able to construct the correct entropy solutions for the critical case.

In summary, the contributions of this paper are as follows:
\begin{enumerate}
    \item This study gives a general framework  of constructing the weak solutions of the hyperbolic conservation laws in non-conservative form or the balance laws with discontinuous perturbations in the flux, for example, the generalized Buckley-Leverett equation with discontinuous porosity \eqref{eqn:ori_GBL}. To the best of our knowledge, PINN or cPINN have not been used for such kind of systems.
    \item The study of critical states—which, as far as we are aware, has not been taken into account in the equations that PINN and cPINN solved—are also covered in this study. To overcome the difficulty that emerges in critical states, we impose the re-scaling process on the unknowns. 
\end{enumerate}

The paper is organized as follows. The review of previous results for the theoretical analysis to the GBL equation is given in Section \ref{sec:GBL}. The review of PINN and cPINN is addressed in Section \ref{sec:cpinn}. Our main results are given in Section
\ref{sec:main}, and the experimental
results are in Section \ref{sec:experiments}, followed by the conclusions in Section
\ref{sec:conclusions}.

\section{Theoretical Results on Riemann problem of the GBL equation}\label{sec:GBL}
The existence and behaviour of the weak solution to the Riemann problem has been proven in the paper \cite{Hong2012}, thus in this section, we briefly review the result.
To convert the equation (\ref{eqn:GBL}) into a $2\times 2$ system of conservation laws, we add $\phi_t = 0$.
Then the Riemann problem of the GBL equation is given by 
\begin{equation}\label{sys:RieGBL}
    \left\lbrace \begin{array}{l}
    U_t + F(U)_x = 0,\\
    U(x,0) = \left \lbrace \begin{array}{l}
            U_L:=(u_L,\phi_L)^T,\quad x<0,\\
            U_R:=(u_R,\phi_R)^T,\quad x>0,
    \end{array}\right.
    \end{array}\right.
\end{equation}
where $U:=(u,\phi)^T$ and $F(U) := (f(u,\phi),0)^T$ and $\phi_L,u_L,\phi_R$ and $u_R$ are constant with $0<u_L\leq \phi_L$, $0<u_L\leq \phi_L$.
We say a $L^1$ function $u$ is a weak function of (\ref{eqn:GBL}) if for any $\psi\in\mathbf{C}_0^{\infty}\times [0,\infty)$, $u$ satisfies 
\begin{equation}\label{dfn:weak}
    \int\int_{t>0} u\psi_t+f(u,x)\psi_xdxdt + \int_{\mathbb{R}} u_0(x)\psi(x,0)dx = 0.
\end{equation}
By direct computation, we know that the eigenvalues and their corresponding right eigenvectors are  
\begin{equation}
    \lambda_0(U) = 0 \quad \textnormal{and} \quad r_0(U) = (f_{\phi},-f_u)^T,
\end{equation}
\begin{equation}
    \lambda_1(U) = f_u(U) = \frac{2M\phi u (\phi-u)}{D^2(u,\phi)} \quad\textnormal{and}\quad r_1(U) = (1,0)^T,
\end{equation}
where $D(u,\phi) = u^2 + M(\phi-u)^2$.
It is clear that the $0-$th characteristic field is linear degenerate due to the fact that $\lambda_0$ is zero.
We have to notice that the first characteristic field is neither genuinely nonlinear nor linear degenerate because $\nabla\lambda_1(U)\cdot r_1(U) = f_{uu}(U)$ has a unique root in $(0,\phi)$.
More precisely,
\begin{equation}\label{eqn:f_uu}
    \nabla\lambda_1(U)\cdot r_1(U) = f_{uu}(U) = \frac{2M\phi}{D^3(u,\phi)}\left[ -u^3 + (\phi-u)(-3u^3+3Mu(\phi-u)+M(\phi-u)^2) \right].
\end{equation}
After straightly forward computation, we know that there exist a unique $m^*$ such that $(u,\phi)$ satisfies $0<u<\phi<m^*u$,
\begin{equation}
f_{uu}<0.
\end{equation}
Similarly, as $(u,\phi)$ satisfies $0<u<m^*u<\phi$,
\begin{equation}
f_{uu}>0.
\end{equation}
For convenience, we define the following two open regions :
\begin{equation}
    \begin{array}{cc}
     & \Omega_{-} =\{(u,\phi)~|~0<u<\phi<m^*u \},  \\
     & \Omega_{+} =\{(u,\phi)~|~0<u<m^*u<\phi \}.
    \end{array}
\end{equation}
The construction of the self-similar type of weak solution to (\ref{sys:RieGBL}) by Lax method requirs to study the element waves for 0-th and 1-th wave fields and their wave curves in $u-\phi$ plane.
Due to the fact that 0-th characteristic field is linear degenerate,
the contact discontinuity connecting two constant states occurs in the solution of (\ref{sys:RieGBL}).
Denoting $\{(u,\phi)\}$ to be the constant states of such contact discontinuity connected to $\{(u_L,\phi_L\}$. 
Then according to the Rankine-Hugoniot condition and the speed of jump is zero, we obtain
\begin{equation}
    f(u,\phi) = f(u_L,\phi_L).
\end{equation}
It is equivalent to
\begin{equation}\label{eqn:jum_0}
    \frac{u^2}{u^2+M(\phi-u)^2} = \frac{u_L^2}{u_L^2+M(\phi_L-u_L)^2}.
\end{equation}
The 0-th shock curve which is also the 0-th rarefaction curve is obtained immediately by solving (\ref{eqn:jum_0}). 
That is, $\{(u,\phi)\}$ satisfy
\begin{equation}\label{eqn:0-th}
u = \frac{u_L}{\phi_L}\phi.
\end{equation}
Before we study the $1-th$ characteristic field,
we point out the fact that if $U_L\in\Omega_{\pm}$ then the curve defined in (\ref{eqn:0-th}) is in $\Omega_{\pm}$.

Next, for the $1-th$ characteristic field $(\lambda_1,r_1)$.
For both rarefaction curve and shock wave, we have $\phi = \phi_R$.
It follows that generalized elementary wave solves
\begin{equation}\label{eqn:single}
    u_t + f(u,\phi_R)_x = 0.
\end{equation}
Firstly, we consider $f_{uu}(u_L,\phi_R)f_{uu}(u_R,\phi_R)>0$. 
The elementary wave is either shock wave or rarefaction wave since $U_L$ and $U_R$ are in the same region $\Omega_{+}$ (or $\Omega_{-}$). 

\begin{figure}
  \centering
  \begin{subfigure}[t]{.23\linewidth}
    \centering\includegraphics[scale=0.45]{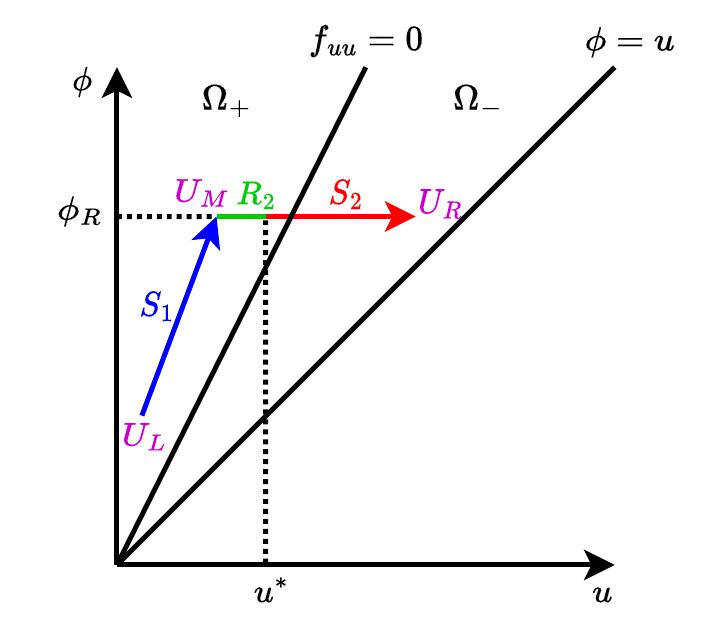}
    \caption{\label{fig:case3}} 
  \end{subfigure}
  \begin{subfigure}[t]{.23\linewidth}
    \centering\includegraphics[scale=0.45]{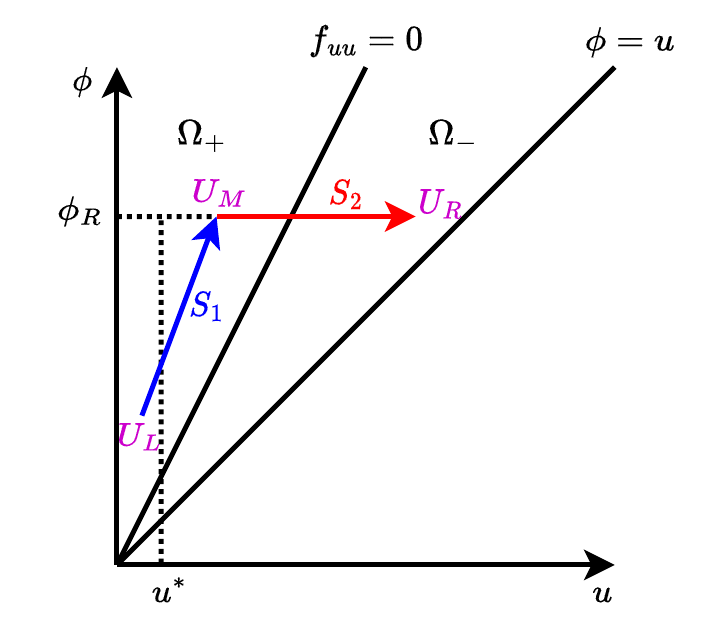}
    \caption{\label{fig:case4}} 
  \end{subfigure}
  \begin{subfigure}[t]{0.23\linewidth}
    \centering\includegraphics[scale=0.45]{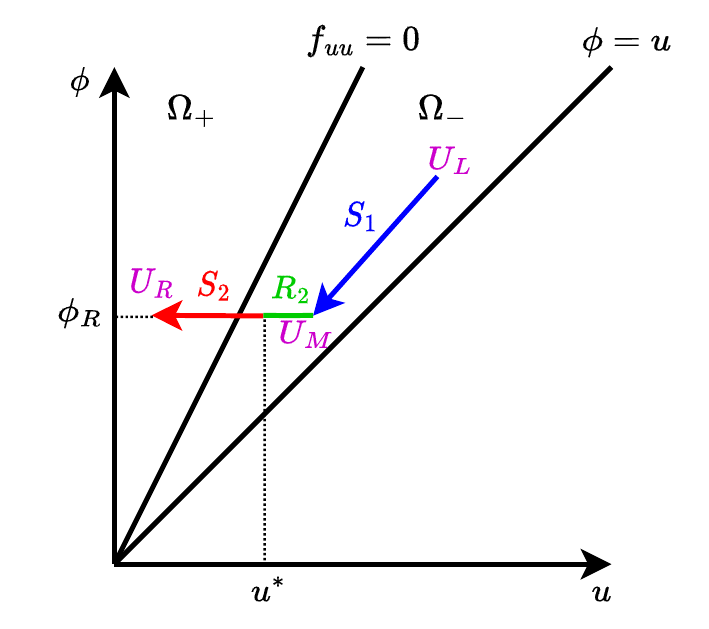}
    \caption{\label{fig:case5}} 
  \end{subfigure}
  \begin{subfigure}[t]{0.23\linewidth}
    \centering\includegraphics[scale=0.45]{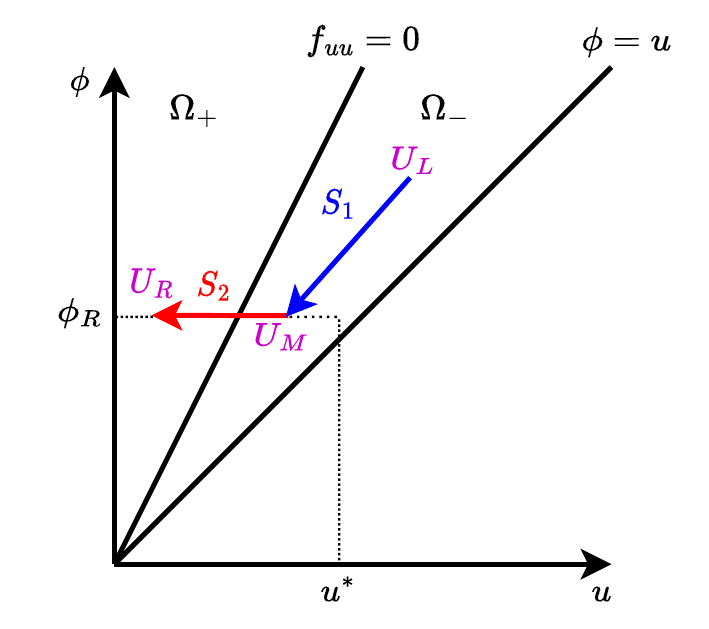}
    \caption{\label{fig:case6}} 
  \end{subfigure}
  \caption{In \ref{fig:case3} and \ref{fig:case4}, $f_{uu}(U_R)<0$ and $f_{uu}(U_L)>0$. The difference is the location of $u^{*}$. In \ref{fig:case3}, $u^{*}>u_{M}$ so that the we have a rarefcation shock wave connecting $U_M$ and $U_R$. But in \ref{fig:case4}, $u^{*}<u_{M}$, in this case $U_M$ and $U_R$ are connected by a shock wave. It is similar for the cases \ref{fig:case5} and \ref{fig:case6}.}\label{fig:case3-6}
\end{figure}

Secondly, if $f_{uu}(u_L,\phi_R)f_{uu}(u_R,\phi_R)<0$.
In this case, $U_L$ and $U_R$ are located in different regions, so the solution wave will cross the region from $\Omega_{+}$ to $\Omega_{-}$ (or reversely).
The difficulty we have in this case is that the flux in equation (\ref{eqn:single}) is neither convex nor concave.
To overcome this difficulty, we define $u^{*}$ by 
\begin{equation}\label{eqn:u-star}
    f_u( u^{*},\phi_R ) = \frac{f(u^{*}, \phi_R) - f(u_R,\phi_R)}{u^{*} - u_R}.
\end{equation}
The existence of $u^{*}$ is stated in \ref{set:u-star}.

According to the Gelfond's construction from Oleinik's work \cite{oleinik1963discontinuous}, we have the following two results.
Given $U_L\in\Omega_{+}$.
If $u_L<u^{*}<u_R$, then the solution wave is a rarefaction-shock wave.
If $u^{*}<u_L<u_R$, then only shock connects two states.
For the case, $U_L\in\Omega_{-}$, the results are symmetric.
In Figure \ref{fig:case3-6}, we show the wave curves for four different cases.

\section{Conservative PINN (cPINN)} \label{sec:cpinn}

In conservative physics-informed neural networks (cPINN), the domain is divided into several sub-domains.
For instance, as the domain is divided into $N_{sd}$ subdomains,
the output for the $n$-th subdomain, is denoted by $\hat{u}_{\theta_n}(z)$ for $n=1,2,\dots, N_{sd}$. Thus, the final output after we stitch back all the subdomains can be written as
\begin{equation}\label{eqn:u_hat_subdms}
    \hat{u}_{\theta}(z) = \bigcup_{n=1}^{N_{sd}} \hat{u}_{\theta_n}(z).
\end{equation}

Let us select training (from initial and boundary), interior, and interface points randomly, denoted as

\begin{equation} \label{eq:points_subdomain}
    \{x^i_{u_n},t^i_{u_n} \}_{i=1}^{N_{u_n}},  \{x^i_{f_n},t^i_{f_n} \}_{i=1}^{N_{f_n}},  \text{and } \{x^i_{I_n},t^i_{I_n} \}_{i=1}^{N_{I_n}}
\end{equation}
respectively, in the $n$-th subdomain. The notations $N_{u_n}, N_{f_n}$, and $N_{I_n}$, respectively, mean the number of points we sample from the training, interior, and interface in the $n$-th subdomain. Moreover, Latin Hypercube Sampling (LHS) \cite{LHS} is used to sample the interior points, whereas random sampling is applied to sample the training and interface points. The cPINN loss function for the $n$-th subdomain is depicted as 

\begin{equation}  \label{eq:loss_cpinn}
    \mathcal{L}(\theta_n) = \omega_{u_n} MSE_{u_n} + \omega_{f_n} MSE_{f_n} + \omega_{I_n} (MSE_{flux_n} + MSE_{avg_n}), \quad
    n = 1,2,\dots,N_{sd},
\end{equation}
where the notations $\omega_{u_n}, \omega_{f_n}$, and $\omega_{I_n}$ are the training, interior, and interface weights, respectively.
Furthermore, the mean square errors (MSE) on the $n$-th subdomain can be described as
\begin{eqnarray}
    MSE_{u_n}    &=&\frac{1}{N_{u_n}} \sum_{i=1}^{N_{u_n}} \left|u_n^i-\hat{u}^i_{\theta_{n}}(x^i_{u_n},t^i_{u_n}) \right|^2 \\
    MSE_{f_n}    &=&\frac{1}{N_{f_n}} \sum_{i=1}^{N_{f_n}} \left|f(x^i_{f_n},t^i_{f_n}) \right|^2 \\
    MSE_{flux_n} &=&\frac{1}{N_{I_n}} \sum_{i=1}^{N_{I_n}} \left| \mathfrak{f}_n (x^i_{I_n},t^i_{I_n}) \cdot \textbf{n} - \mathfrak{f}_{n^+} (x^i_{I_n},t^i_{I_n}) \cdot \textbf{n} \right|^2 \\
    MSE_{avg_n}  &=&\frac{1}{N_{I_n}} \sum_{i=1}^{N_{I_n}} \left| \hat{u}^i_{\theta_{n}}(x^i_{I_n},t^i_{I_n}) - \big\{ \big\{\hat{u}^i_{\theta_{n}}(x^i_{I_n},t^i_{I_n})\big\} \big\} \right|^2
\end{eqnarray}
where the parameters of the $n$-th subdomain are represented by the subscript $\theta_n$. The symbol $f$ denotes the governing equation's residual; as computing the residual necessitates employing the derivatives of the independent variables based on the governing equation, automatic differentiation (AD) \cite{AD} is required. Additionally, $\mathfrak{f}_n$ represents the flux in the $n$-th subdomain. The adjacent subdomains are indicated by the superscript $+$. Moreover, the average $\hat{u}^i_{\theta_{n}}$ value throughout the shared interface across the subdomains is indicated as
\begin{equation} \label{eq:avg}
    \hat{u}_{avg} = \big\{ \big\{\hat{u}^i_{\theta_{n}}(x^i_{I_n},t^i_{I_n})\big\} \big\} \triangleq \frac{\hat{u}^i_{\theta_{n}} + \hat{u}^i_{\theta_{n^+}}}{2}.
\end{equation}


\section{cPINN for Generalized Buckley-Leverett Equations}
\label{sec:main}

\begin{figure}
    \centering
  \includegraphics[scale=0.55]{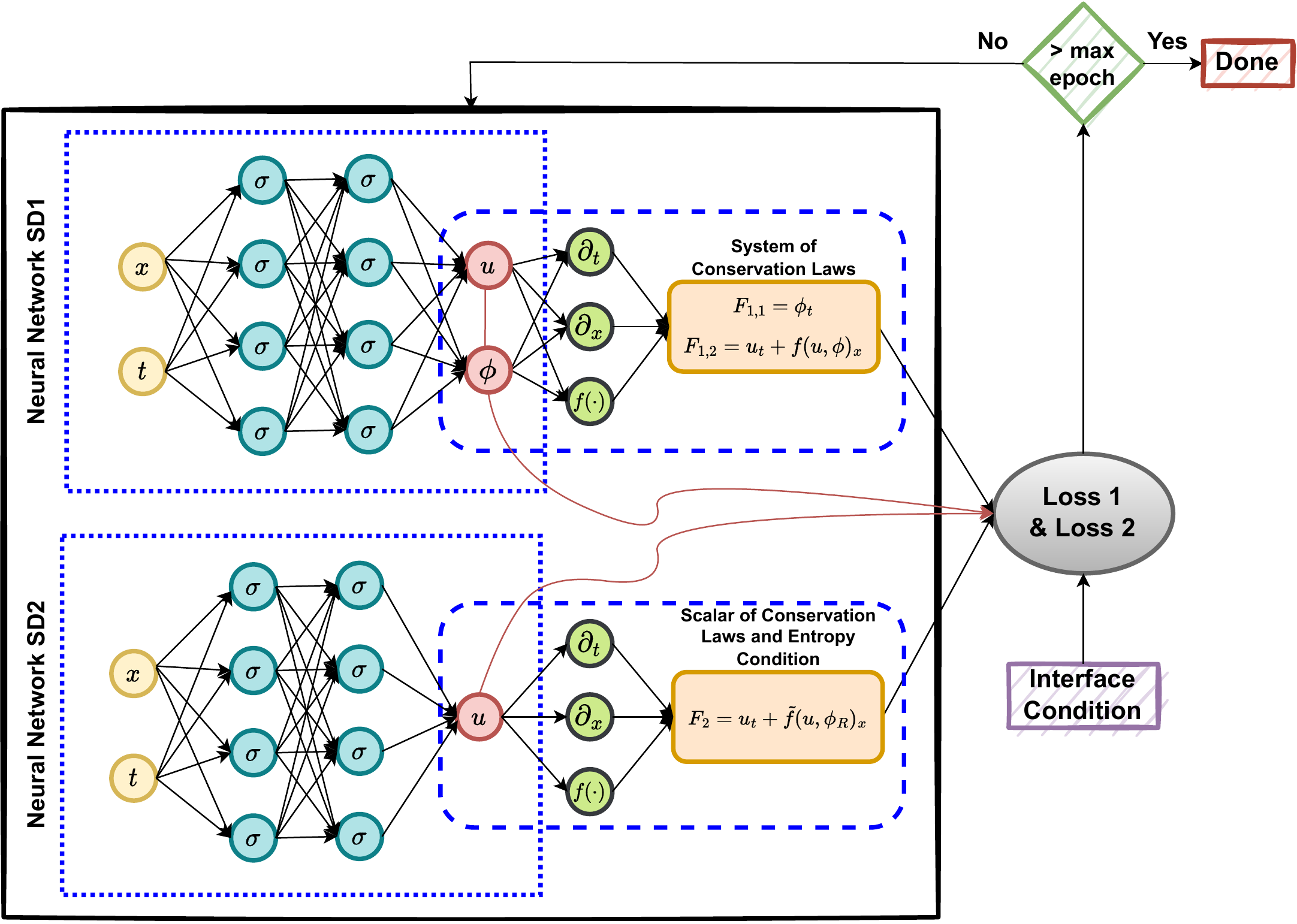}
  \caption{The schematic of cPINN to solve generalized Buckley-Leverett equation. The domain is divided into two subdomains by an interface: the first subdomain solves a system, while the second subdomain handles a scalar equation with an entropy condition.}
  \label{fig:flowchart}
\end{figure}

The domain was divided into two subdomains, with subdomain 1 (SD1) handling the system case and subdomain 2 (SD2) solving the scalar case. As a result of dealing with the system case in subdomain 1, the neural network outputs are $\phi$ and $u$. The output of subdomain 2, however, is merely $u$, with the variable $\phi$ fixed at $\phi_R$. This implementation adheres to the theoretical procedure for determining the generalized Buckley-Leverett solution (explained in section \ref{sec:GBL}). In addition, an extra constraint, such as the Oleinik entropy condition, must be enforced. The Oleinik entropy condition is essential when dealing with a solution that involves  shock, the entropy condition is therefore applied in the second subdomain. To illustrate, Figure \ref{fig:flowchart} depicts the schematic representation of cPINN used to solve the generalized Buckley-Leverett equation.

In this study, we investigate a variety of examples involving conservative and non-conservative forms, as well as non-critical and critical states. As a result, each type of case affects the loss function.

\subsection{Conservative Form} \label{subsec:conservative_form}
The Conservative Form is based on the equation (\ref{sys:GBL}).
\subsubsection{Non-Critical States} \label{subsubsec:non_vacuum_conservative}

In order to incorporate cPINN into our generalized Buckley-Leverett equation (\ref{sys:GBL}), we adjust the cPINN loss function (\ref{eq:loss_cpinn}) in subdomain 1 into the following equation.

\begin{equation} \label{eq:c_loss1}
    \textnormal{loss}_1 = \omega_{u_1} MSE_{u_1} + \omega_{f_1} MSE_{f_1} + \omega_{I_1} (MSE_{flux_1} + MSE_{avg_1})
\end{equation}
where
\begin{align}
    MSE_{u_1} = &
  \!\begin{aligned}[t] 
    \frac{1}{N_{u_1}} \sum_{i=1}^{N_{u_1}} \left|u^i - \hat{u}_{\theta_1}^i (x^i_{u_1},t^i_{u_1}) \right|^2 + \frac{1}{N_{u_1}} \sum_{i=1}^{N_{u_1}} \left|\phi^i - \hat{\phi}_{\theta_1}^i (x^i_{u_1},t^i_{u_1}) \right|^2 
  \end{aligned}\\
  MSE_{f_1} = &
  \!\begin{aligned}[t]
     &\frac{1}{N_{f_1}} \sum_{i=1}^{N_{f_1}} \left| \big(\hat{\phi}_{\theta_1}^i (x^i_{f_1},t^i_{f_1}) \big)_t \right|^2 \\ &+ \frac{1}{N_{f_1}} \sum_{i=1}^{N_{f_1}} \left| \big(\hat{u}_{\theta_1}^i (x^i_{f_1},t^i_{f_1}) \big)_t + f \big( \hat{u}_{\theta_1}^i (x^i_{f_1},t^i_{f_1}), \hat{\phi}_{\theta_1}^i  (x^i_{f_1},t^i_{f_1}) \big)_x \right|^2 \\
  \end{aligned}\\
  MSE_{flux_1} = &
    \!\begin{aligned}[t]
     \frac{1}{N_{I_1}} \sum_{i=1}^{N_{I_1}} \left| f \big( \hat{u}^i_{\theta_1} (x^i_{I_1},t^i_{I_1}),\hat{\phi}^i_{\theta_1} (x^i_{I_1},t^i_{I_1}) \big) -  f \big( \hat{u}^i_{\theta_2} (x^i_{I_1},t^i_{I_1}), \phi_R \big) \right|^2 \\
  \end{aligned}\\
  MSE_{avg_1} =&
    \!\begin{aligned}[t]
     &\frac{1}{N_{I_1}} \sum_{i=1}^{N_{I_1}} \left| \hat{u}^i_{\theta_1} (x^i_{I_1},t^i_{I_1}) -\big\{ \big\{\hat{u}^i_{\theta_1} (x^i_{I_1},t^i_{I_1})\big\} \big\} \right|^2 \\ & + \frac{1}{N_{I_1}} \sum_{i=1}^{N_{I_1}} \left| \hat{\phi}^i_{\theta_1} (x^i_{I_1},t^i_{I_1}) - \big\{ \big\{\hat{\phi}^i_{\theta_1}(x^i_{I_1},t^i_{I_1})\big\} \big\}  \right|^2
  \end{aligned}
\end{align}
where the notations $(x^i_{u_1},t^i_{u_1})$,  $(x^i_{f_1},t^i_{f_1})$,  and $(x^i_{I_1},t^i_{I_1})$ follows the equation \eqref{eq:points_subdomain}, they indicate the respective training, interior, and interface points are randomly selected from subdomain 1. Moreover, the outputs of neural network in SD1 are $\hat{u}_{\theta_1}^i$ and $\hat{\phi}_{\theta_1}^i$, and the initial condition provides the value of $u^i$ and $\phi^i$. In addition, the residual $f$ follows the equation \eqref{eqn:flux}. Also, the notation $\{ \{ \cdot \} \}$ is represented in equation (\ref{eq:avg}). The parameters $\theta_1$ are trained to minimize loss 1 (equation (\ref{eq:c_loss1})) and belong to subdomain 1. Meanwhile, the parameters $\theta_2$ are associated with subdomain 2.  

The loss function for subdomain 2 is provided below.
\begin{equation} \label{eq:c_loss2}
    loss_2 = \omega_{u_2} MSE_{u_2} + \omega_{f_2} MSE_{f_2} + \omega_{I_2} (MSE_{flux_2} + MSE_{avg_2})
\end{equation}
where
\begin{align}
    MSE_{u_2} = &
  \!\begin{aligned}[t] 
    \frac{1}{N_{u_2}} \sum_{i=1}^{N_{\theta_2}} \left|u_{\theta_2}^i (x^i_{u_2},t^i_{u_2}) - \hat{u}_{\theta_2}^i (x^i_{u_2},t^i_{u_2}) \right|^2
  \end{aligned}\\
  MSE_{f_2} = &
  \!\begin{aligned}[t]
     \frac{1}{N_{f_2}} \sum_{i=1}^{N_{f_2}} \left| \big(\hat{u}_{\theta_2}^i (x^i_{f_2},t^i_{f_2}) \big)_t + \tilde{f} \big( \hat{u}_{\theta_2}^i (x^i_{f_2},t^i_{f_2}), \phi_R  \big)_x \right|^2
  \end{aligned}\\
  MSE_{flux_2} = &
    \!\begin{aligned}[t]
     \frac{1}{N_{I_1}} \sum_{i=1}^{N_{I_1}} \left| f \big( \hat{u}^i_{\theta_1} (x^i_{I_1},t^i_{I_1}),\hat{\phi}^i_{\theta_1} (x^i_{I_1},t^i_{I_1}) \big) -  f \big( \hat{u}^i_{\theta_2} (x^i_{I_1},t^i_{I_1}), \phi_R \big) \right|^2 \\
  \end{aligned}\\
  MSE_{avg_2} =&
    \!\begin{aligned}[t]
     &\frac{1}{N_{I_1}} \sum_{i=1}^{N_{I_1}} \left| \hat{u}^i_{\theta_2} (x^i_{I_1},t^i_{I_1}) -\big\{ \big\{\hat{u}^i_{\theta_1} (x^i_{I_1},t^i_{I_1})\big\} \big\} \right|^2 \\ &+ \frac{1}{N_{I_1}} \sum_{i=1}^{N_{I_1}} \left| \phi_R - \big\{ \big\{\hat{\phi}^i_{\theta_1}(x^i_{I_1},t^i_{I_1})\big\} \big\}  \right|^2
  \end{aligned}
\end{align}
The notation's interpretation is similar to the one in loss 1 \eqref{eq:c_loss1}. Since we are aware that cPINN cannot accommodate moving shocks in solutions, we need an extra constraint to make the solution more reasonable; thus, the Oleinik entropy condition is incorporated in the loss function. The Oleinik entropy condition is denoted by $\tilde{f}(u,\phi)$, where

\begin{equation}\label{eq:oleinik_ec}
    \tilde{f}(u,\phi) = \left \lbrace \begin{array}{ll}
            \tilde{f}_1(u,\phi), & u_M > u_R \textnormal{ and } u_M > u^*\\
            su,  & u_M > u_R \textnormal{ and } u_M < u^*\\
            \tilde{f}_1(u,\phi), & u_M < u_R \textnormal{ and } u_M < u^*\\
            su,  & u_M < u_R \textnormal{ and } u_M > u^*
    \end{array}\right.
\end{equation}

and
\begin{equation}
\tilde{f}_1(u,\phi) = \left \lbrace \begin{array}{ll}
            su,  &\textnormal{for shock} \\
            f(u,\phi), \quad &\textnormal{for rarefaction}
    \end{array}\right.
\end{equation}

and $s$ is the speed determined by the Rankine-Hugoniot condition.

\subsubsection{Critical States} \label{subsubsec:vacuum_conservative}
For the critical states, we must rescale equation (\ref{sys:GBL}) by applying the relationship 
\begin{equation}\label{eqn:rescale_4}
 \phi=\delta_1 \bar{\phi}\quad \text{and} \quad u = \delta_2 \bar{u}   
\end{equation}
resulting in the following equation.
\begin{equation}\label{eq:c_rescale_form}
    \left \lbrace \begin{array}{ll}
            \bar{\phi}_t &= 0 \\
            \delta_2 \bar{u}_t + \left( \frac{\bar{u}^2}{\bar{u}^2 + M \left(\frac{\delta_1}{\delta_2} \bar{\phi} - \bar{u} \right)^2} \right)_x &= 0
    \end{array}\right.
\end{equation}
 When the critical state takes place in $U_L$, all that needs to be adjusted is the loss function in subdomain 1; as a result, $MSE_{f_1}$ in loss 1 (equation (\ref{eq:c_loss1})) needs to be modified based on equation (\ref{eq:c_rescale_form}), resulting in the equation shown below.

\begin{equation} \label{eq:C_vacuum_ul}
    MSE_{f_1} = \frac{1}{N_{f_1}} \sum_{i=1}^{N_{f_1}} \left| \big(\hat{\bar{\phi}}_1^i \big)_t \right|^2 + \frac{1}{N_{f_1}} \sum_{i=1}^{N_{f_1}} \left| \delta_2 \big(\hat{\bar{u}}_1^i \big)_t + \bar{f} \big( \hat{\bar{u}}_1^i, \hat{\bar{\phi}}_1^i  \big)_x \right|^2,
\end{equation}
where
\begin{equation}
    \bar{f}(u,\phi) = \frac{u^2}{u^2 + M \big(\frac{\delta_1}{\delta_2} \phi - u \big)^2}.
\end{equation}
The notations $\hat{\bar{u}}_1^i$ and $\hat{\bar{\phi}}_1^i$ refer to the output of the neural network SD1 in Figure \ref{fig:flowchart}  from the randomly chosen interior points, and afterward $\hat{\phi}_1^i$ and $\hat{u}_1^i$ are obtained by equation (\ref{eqn:rescale_4}).

When the critical occurs in $U_R$, we only need to modify the loss function in subdomain 2 (equation (\ref{eq:c_loss2})). Therefore, only $MSE_{f2}$ needs to be altered, leading to the following equation.

\begin{equation} \label{eq:C_vacuum_ur}
    MSE_{f_2} =  \frac{1}{N_{f_2}} \sum_{i=1}^{N_{f_2}} \left| \big(\hat{\bar{u}}_2^i \big)_t + \tilde{f} \big( \hat{\bar{u}}_2^i, \bar{\phi}_R, \delta_1, \delta_2  \big)_x \right|^2,
\end{equation}
where 

\begin{equation}\label{eq:oleinik_ec_rescale}
    \tilde{f}(u,\phi,\delta_1, \delta_2) = \left \lbrace \begin{array}{ll}
            \tilde{f}_1(u,\phi,\delta_1, \delta_2), & u_M > u_R \textnormal{ and } u_M > u^*\\
            (s/\delta_2)u,  & u_M > u_R \textnormal{ and } u_M < u^*\\
            \tilde{f}_1(u,\phi,\delta_1, \delta_2), & u_M < u_R \textnormal{ and } u_M < u^*\\
            (s/\delta_2)u,  & u_M < u_R \textnormal{ and } u_M > u^*
    \end{array}\right.
\end{equation}
and
\[
\tilde{f}_1(u,\phi,\delta_1, \delta_2) = 
    \left \lbrace \begin{array}{ll}
            \frac{s}{\delta_2} u, & \textnormal{ for shock}\\
            \frac{1}{\delta_2} \frac{u^2}{u^2 + M \big( \frac{\delta_1}{\delta_2} \phi - u \big)^2}, & \textnormal{ for rarefaction}
    \end{array}\right.
\]
where $\hat{\bar{u}}_2^i$ is the notation after feeding the input from the residual points and implementing $\hat{u}_2 = \delta_2 \hat{\bar{u}}_2$. Note that the traveling shock's speed has been altered from the original Oleinik entropy condition  \eqref{eq:oleinik_ec} in the rescaling case.

\subsection{Non-Conservative Form}\label{subsec:non_conservative_form}
We obtained the non-conservative form of the Generalized Buckley-Leverett equation below by applying $u = \phi \tilde{u}$ to the conservative form (equation (\ref{sys:GBL})).

\begin{equation} \label{eq:nc_form}
    \left \lbrace \begin{array}{ll}
            \phi_t &= 0 \\
            \phi \tilde{u}_t + g(\tilde{u})_x &= 0
    \end{array}\right.
\end{equation}
where $g(\tilde{u})= \frac{\tilde{u}^2}{\tilde{u}^2 +M(1-\tilde{u})^2}$.

\subsubsection{Non-Critical States}\label{subsubsec:non_vacuum_nonconservative} 
The modification of the loss function of the non-conservative form is relatively straightforward, by utilizing equation (\ref{eq:nc_form}) as the governed equation.

\subsubsection{Critical States}\label{subsubsec:vacuum_nonconservative}
The non-conservative form of the rescaling in a critical state can be altered in a straightforward manner, similarly to the conservative form, by employing the relationships $\phi = \delta_1 \bar{\phi}$ and $u = \delta_2 \bar{u}$ solely in the interior point. As a consequence, the underlying equation in the non-conservative for rescaling in a critical state is as follows.

\begin{equation} \label{eq:nc_rescale_form}
    \left \lbrace \begin{array}{ll}
            \bar{\phi}_t &= 0 \\
            \delta_1 \delta_2 \bar{\phi} \bar{u}_t + \left( \frac{\bar{u}^2}{\bar{u}^2 + M \left(\frac{1}{\delta_2} \bar{\phi} - \bar{u} \right)^2} \right)_x &= 0
    \end{array}\right.
\end{equation}

\section{Numerical results}
\label{sec:experiments}
In this section, we detail the results of numerous numerical experiments. The cases are classified into two categories: conservative and non-conservative. Equation (\ref{sys:GBL}) gives the generalized Buckley-Leverett equation's conservative form, while equation \eqref{eqn:ori_GBL} gives its non-conservative form. We also conducted studies on non-critical and critical states in each category.  Each of the numerical experiment was performed out three times, hence the average $L_2$ norm given in this paper is the average of three runs. In addition, some comparisons with WENO5 are offered in section \ref{set:comp_weno5}.  The detail of the experimental settings is provided in \ref{experiment_detail_appendix}.

\subsection{Non-Critical States}\label{subsubsec:conservative_non_vacuum_experiment}
We implement the loss in subdomain 1 (loss 1) and the loss in subdomain 2 (loss 2) for the conservative form in the non-critical cases by using the equations (\ref{eq:c_loss1}) and (\ref{eq:c_loss2}), respectively. Furthermore, all of the non-critical cases share the same domain, i.e., $0 \leq t \leq 3$ and $-1 \leq x \leq 10$. The general initial conditions can be expressed as follows. 

\[ U(x,0) = 
    \left \lbrace \begin{array}{ll}
            U_L, & \textnormal{if}\ x<0 \\
            U_R, & \textnormal{else}
    \end{array}\right.
\]

\subsubsection{Case 1} \label{case1}

Following are the initial conditions for the first case in both conservative and non-conservative forms:
\begin{equation} \label{eq:iv_case1_c}
    U_L = \begin{pmatrix}
    u_L \\ 
    \phi_L
    \end{pmatrix} = 
    \begin{pmatrix}
    0.6 \\ 
    0.7
    \end{pmatrix} \quad \text{ and } \quad
    U_R = \begin{pmatrix}
    u_R \\ 
    \phi_R
    \end{pmatrix} = 
    \begin{pmatrix}
    0.3 \\ 
    0.6
    \end{pmatrix},
\end{equation}

\begin{equation} \label{eq:iv_case1_nc}
    \Tilde{U}_L = \begin{pmatrix}
    \Tilde{u}_L \\ 
    \phi_L
    \end{pmatrix} = 
    \begin{pmatrix}
    6/7 \\ 
    0.7
    \end{pmatrix} \quad \text{ and } \quad
    \Tilde{U}_R = \begin{pmatrix}
    \Tilde{u}_R \\ 
    \phi_R
    \end{pmatrix} = 
    \begin{pmatrix}
    0.5 \\ 
    0.6
    \end{pmatrix},
\end{equation}
respectively.

\begin{figure}
    \centering
    \includegraphics[scale=0.45]{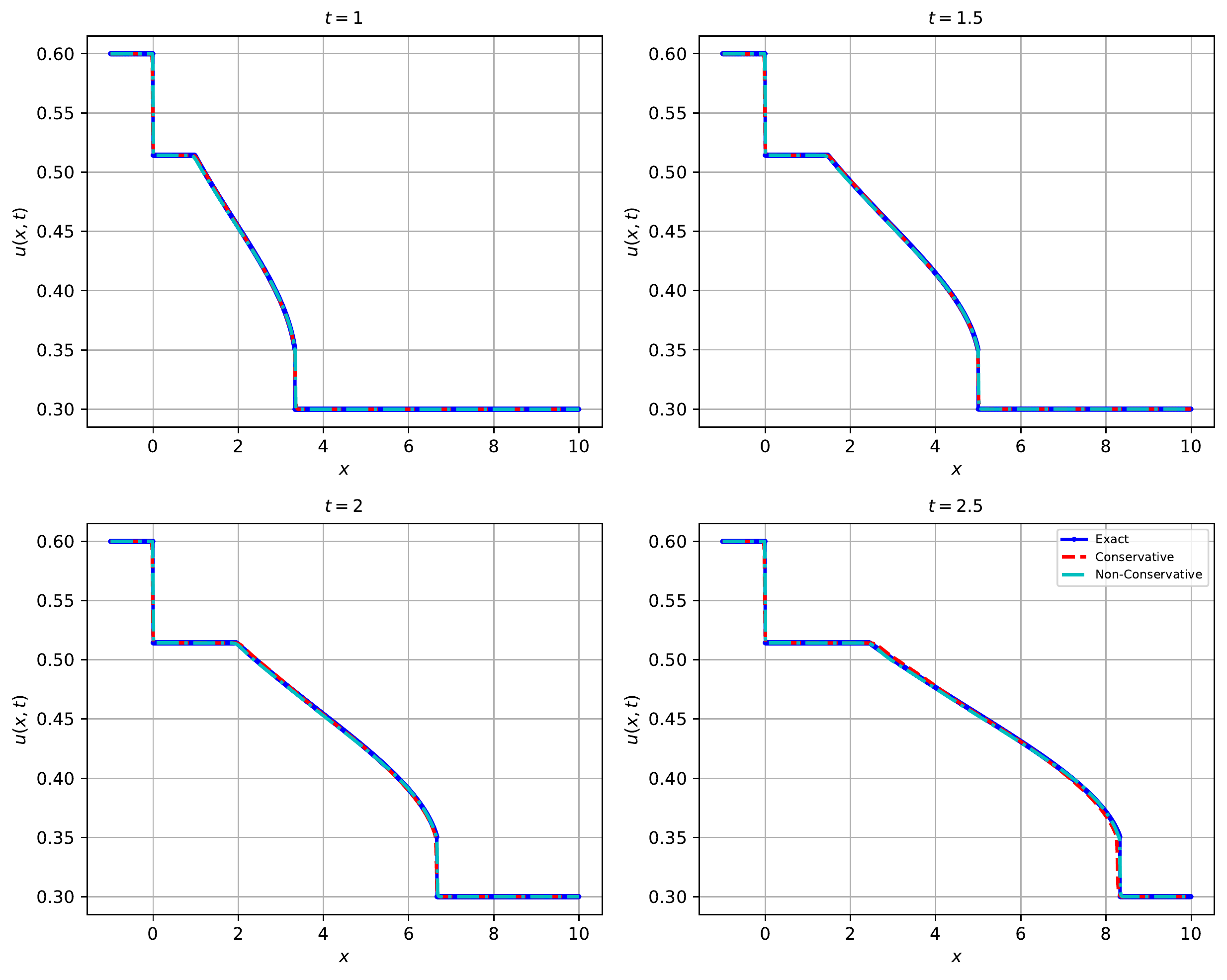}
    \caption{Comparison of the exact and predicted solution in conservative and non-conservative forms at various time-step for the GBL equation with initial value \eqref{eq:iv_case1_c} and \eqref{eq:iv_case1_nc}. The average $L_2$ norm is $8.96 \cdot 10^{-3}$ for the conservative form, and $6.05 \cdot 10^{-3}$ for the non-conservative form.}
    \label{fig:case1}
\end{figure}

The theoretical illustration is available in Figure \ref{fig:case5}. By the derivation of some algebraic calculations in equation (\ref{eqn:0-th}), we obtain  $u_M \approx 0.51$. The result of the conservative and the non-conservative form is depicted in Figure \ref{fig:case1}, with the average relative $L_2$ norm after three repetitions of the experiment being $8.96 \cdot 10^{-3}$ for the conservative, and $6.05 \cdot 10^{-3}$ for the non-conservative form. As demonstrated, both with initial condition in conservative \eqref{eq:iv_case1_c} and non-conservative \eqref{eq:iv_case1_nc}, we can successfully acquire the solution rarefaction and shock waves.

\subsubsection{Case 2} \label{case2}
The following is the initial condition:

\begin{equation} \label{eq:iv_case2_c}
    U_L = \begin{pmatrix}
    u_L \\ 
    \phi_L
    \end{pmatrix} = 
    \begin{pmatrix}
    0.45 \\ 
    0.8
    \end{pmatrix} \quad \text{ and } \quad
    U_R = \begin{pmatrix}
    u_R \\ 
    \phi_R
    \end{pmatrix} = 
    \begin{pmatrix}
    0.3 \\ 
    0.6
    \end{pmatrix}
\end{equation}
for the conservative form, and

\begin{equation} \label{eq:iv_case2_nc}
    \Tilde{U}_L = \begin{pmatrix}
    \Tilde{u}_L \\ 
    \phi_L
    \end{pmatrix} = 
    \begin{pmatrix}
    0.5625 \\ 
    0.8
    \end{pmatrix} \quad \text{ and } \quad
    \Tilde{U}_R = \begin{pmatrix}
    \Tilde{u}_R \\ 
    \phi_R
    \end{pmatrix} = 
    \begin{pmatrix}
    0.5 \\ 
    0.6
    \end{pmatrix}
\end{equation}
for the non-conservative form.

\begin{figure}
    \centering
    \includegraphics[scale=0.45]{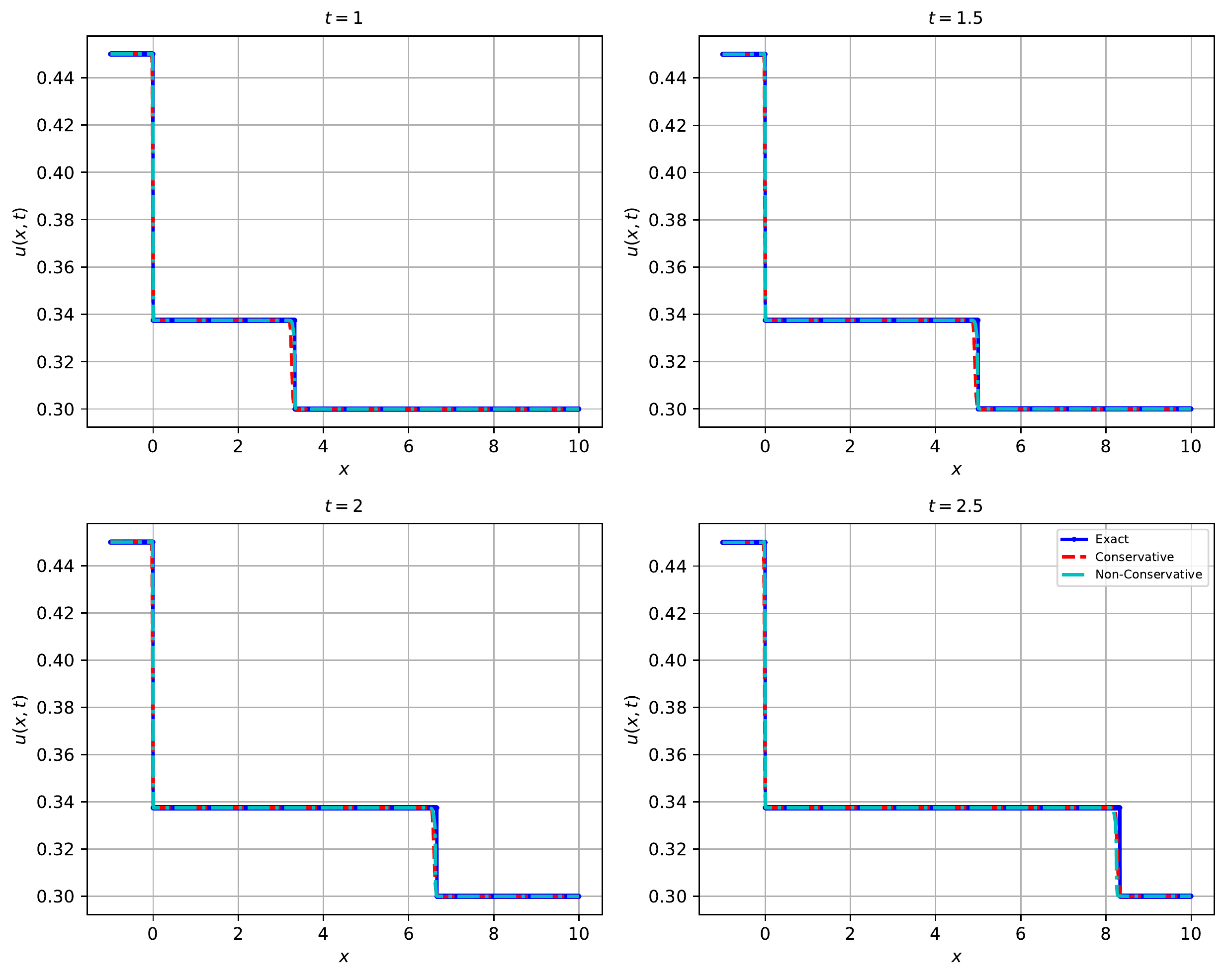}
    \caption{Comparison of the exact and predicted solution in conservative and non-conservative forms at various time-step for the GBL equation with initial value \eqref{eq:iv_case2_c} and \eqref{eq:iv_case2_nc}. The average $L_2$ norm are $1.11 \cdot 10^{-2}$ and $8.82 \cdot 10^{-3}$ for the conservative and non-conservative forms, respectively.}
    \label{fig:case2}
\end{figure}

Figure \ref{fig:case6} offers the theoretical illustration. Similar to Section \ref{case1}, we apply equation (\ref{eqn:0-th}) to obtain $u_M = 0.3375$. The exact and predicted solution of cPINN in conservative and non-conservative forms are presented in Figure \ref{fig:case2}. The average relative $L_2$ norm is $1.11 \cdot 10^{-2}$ and $8.82 \cdot 10^{-3}$, respectively for the conservative and non-conservative form. We notice that cPINN can accurately capture the solution in both conservative and non-conservative forms. In contrast to Case \ref{case1}, the current case's solution consists only of shock waves.

\subsection{Critical States}\label{subsubsec:conservative_vacuum_experiment}

\subsubsection{Case 3a} \label{case3a} This case has a domain of $0 \leq t \leq 3$ and $-1 \leq x \leq 10$. The equations (\ref{eq:c_loss1}) and (\ref{eq:c_loss2}) are used for loss 1 and 2, respectively, where the loss function is equivalent to that of the non-critical state.
The following is the initial condition for the conservative form of the critical state that occurs in $U_L$:

\begin{equation} \label{eq:iv_case3_c}
    U_L = \begin{pmatrix}
    u_L \\ 
    \phi_L
    \end{pmatrix} = 
    \begin{pmatrix}
    2 \cdot 10^{-4} \\ 
    0.1
    \end{pmatrix} \quad \text{ and } \quad
    U_R = \begin{pmatrix}
    u_R \\ 
    \phi_R
    \end{pmatrix} = 
    \begin{pmatrix}
    0.35 \\ 
    0.5
    \end{pmatrix}.
\end{equation}

Furthermore, the initial condition for the non-conservative form where the critical state is in $\Tilde{U}_L$ is as follows:

\begin{equation} \label{eq:iv_case3_nc}
    \Tilde{U}_L = \begin{pmatrix}
    \Tilde{u}_L \\ 
    \phi_L
    \end{pmatrix} = 
    \begin{pmatrix}
    2 \cdot 10^{-3} \\ 
    0.1
    \end{pmatrix} \quad \text{ and } \quad
    \Tilde{U}_R = \begin{pmatrix}
    \Tilde{u}_R \\ 
    \phi_R
    \end{pmatrix} = 
    \begin{pmatrix}
    0.7 \\ 
    0.5
    \end{pmatrix}
\end{equation}

\begin{figure}[ht]
    \centering\includegraphics[scale=0.45]{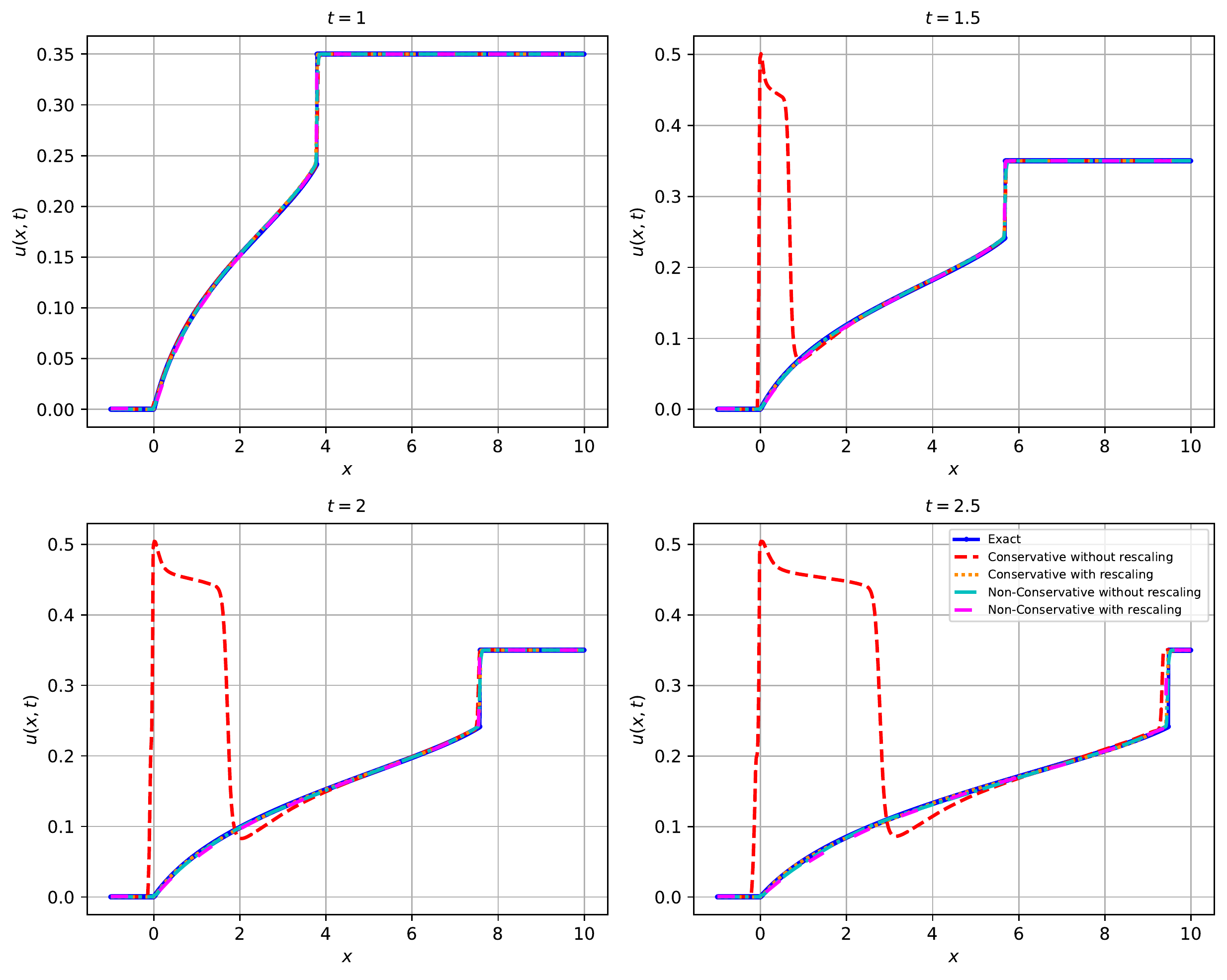}
    \caption{Comparisons are made between the exact and cPINN solutions for the GBL equation's critical case in $U_L$ in both conservative and non-conservative forms, also with and without rescaling. As one can observe, in conservative form, cPINN was unable to effectively handle the critical case (red dashed line); however, after applying the rescaling technique, the outcome is excellent (orange dots line). The average relative $L_2$ norm before rescaling is $3.33 \cdot 10^{-1}$, whereas after rescaling, the relative $L_2$ norm is $1.73 \cdot 10^{-2}$. Unexpectedly, the non-conservative form of the cPINN performs remarkably well when handling the critical case in $U_L$ (cyan dashed line). We also run the case where we implement the rescaling on the non-conservative form, and it works as expected (pink dashed line). Without and with rescaling, the average $L_2$ norm of the non-conservative form is $1.54 \cdot 10^{-3}$ and $2.21 \cdot 10^{-3}$, respectively. Based on the average $L_2$ norm, it is not required to use the rescaling technique on the non-conservative form, in this case, as it will not enhance the performance of cPINN.}
 \label{fig:case3ab}
\end{figure}

The theoretical illustration is available in Figure \ref{fig:case3}. The outcome of the conservative form is shown in a red dashed line in Figure \ref{fig:case3ab}, with the average relative $L_2$ norm is $3.33 \cdot 10^{-1}$. As one can see, the critical state is where cPINN in conservative form fails to function well. As a result, in the following case, we use the rescaling technique to address the problem in the conservative form. Surprisingly, in the critical state in $\Tilde{U}_L$, cPINN in a non-conservative form (marked by the cyan dashed line in Figure \ref{fig:case3ab}) works well.
The average $L_2$ norm for the non-conservative form is $1.54 \cdot 10^{-2}$.

\subsubsection{Case 3b} \label{case3b}
This case has the same domain as Section \ref{case3a}, and equations \eqref{eq:iv_case3_c} and \eqref{eq:iv_case3_nc} provide the initial condition in conservative and non-conservative forms, respectively. For the conservative form, we employ the rescaling technique to solve the issue  when a critical state emerges in $U_L$. We simply need to rescale in the first subdomain (loss 1) and alter the interior parts. Loss 1 (equation (\ref{eq:c_loss1})) with the $MSE_{f_1}$ (equation (\ref{eq:C_vacuum_ul})) where $\delta_1= 10^{-2}$ and $\delta_2=10^{-4}$ is therefore considered. We observe that rescaling improves cPINN's performance in the critical state as shown in orange-dots line in Figure \ref{fig:case3ab}. After three repetitions, we obtain the average relative $L_2$ norm for the conservative form as $1.73 \cdot 10^{-2}$.

As shown in Section \ref{case3a}, the non-conservative form without rescaling has no trouble approximating the solution of the critical state in $U_L$. However, we continue to implement the non-conservative form with rescaling experiment for research purposes. We used $\delta_1= 10^{-2}$ and $\delta_2=10^{-3}$ as rescaling parameters to implement the rescaling method. Thus, the result can be seen as a pink-dashed line in Figure \ref{fig:case3ab}. As we suspect, non-conservative form with rescaling can handle the critical case in $U_L$ quite effectively. Furthermore, the relative $L_2$ norm of the non-conservative form with rescaling is $2.21 \cdot 10^{-2}$, which is greater than the one without rescaling, indicating that, in this case, rescaling in the non-conservative form is unnecessary because it will not improve cPINN performance.

\subsubsection{Case 4a} \label{case4a}
\begin{figure}
    \centering\includegraphics[scale=0.45]{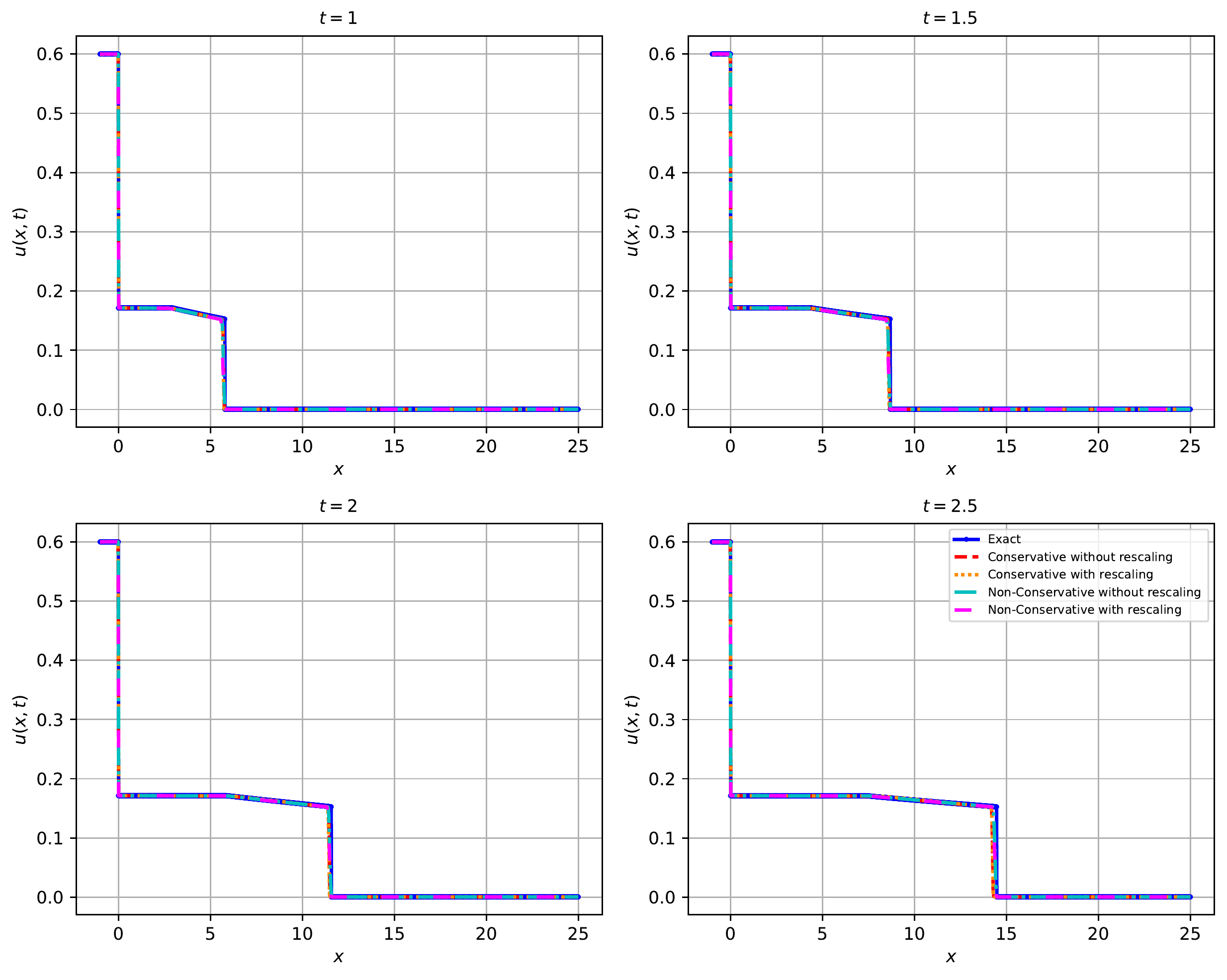}
    \caption{Comparison of the exact solution and the cPINN solution in conservative and non-conservative forms, as well as without and with rescaling applied to the GBL equation's critical state in $U_R$ are presented. In general, cPINN can perform pretty well when the critical state is in $U_R$, regardless of conservative or non-conservative, without or with rescaling. The relative $L_2$ norm for the conservative form are $7.2 \cdot 10^{-2}$ and $8.82 \cdot 10^{-2}$, for without and with rescaling, respectively. Additionally, for the non-conservative form, the relative $L_2$ norm are $6.09 \cdot 10^{-2}$ and $5.99 \cdot 10^{-2}$ for without and with rescaling, respectively. Based on the $L_2$ norm, the cPINN performance in the non-conservative form is slightly enhanced by rescaling.
}
 \label{fig:case4ab}
\end{figure}

For this case, we select the initial condition of the critical state occurs in $U_R$ in the conservative form as follows:

\begin{equation} \label{eq:iv_case4_c}
    U_L = \begin{pmatrix}
    u_L \\ 
    \phi_L
    \end{pmatrix} = 
    \begin{pmatrix}
    0.6 \\ 
    0.7
    \end{pmatrix} \quad \text{ and } \quad
    U_R = \begin{pmatrix}
    u_R \\ 
    \phi_R
    \end{pmatrix} = 
    \begin{pmatrix}
    4 \cdot 10^{-4} \\ 
    0.2
    \end{pmatrix},
\end{equation}
while the subsequent is the initial condition for the non-conservative form:
\begin{equation} \label{eq:iv_case4_nc}
    \Tilde{U}_L = \begin{pmatrix}
    \Tilde{u}_L \\ 
    \phi_L
    \end{pmatrix} = 
    \begin{pmatrix}
    6/7 \\ 
    0.7
    \end{pmatrix} \quad \text{ and } \quad
    \Tilde{U}_R = \begin{pmatrix}
    \Tilde{u}_R \\ 
    \phi_R
    \end{pmatrix} = 
    \begin{pmatrix}
    2 \cdot 10^{-3} \\ 
    0.2
    \end{pmatrix},
\end{equation}
where their domain is $0 \leq t \leq 3$ and $-1 \leq x \leq 25$. The theoretical illustration is available in Figure \ref{fig:case5}. The loss function that is employed in the conservative form is provided in the equations (\ref{eq:c_loss1}) and (\ref{eq:c_loss2}), respectively, in SD1 and SD2. Figure \ref{fig:case4ab} shows the results of the conservative and non-conservative forms without rescaling in red and cyan dashed lines, respectively.
Based on Figure \ref{fig:case4ab}, both perform quite satisfactorily, with the conservative form's relative $L_2$ norm being $7.2 \cdot 10^{-2}$ and the non-conservative form's relative $L_2$ norm being $6.09 \cdot 10^{-2}$. 

\subsubsection{Case 4b} \label{case4b}
Given the initial conditions in equation \eqref{eq:iv_case4_c} and \eqref{eq:iv_case4_nc}, also the same domain as Case \ref{case4a}, we perform the rescaling technique. Take into account, for the conservative form, we simply need to adjust the loss in the second subdomain since the critical state currently exists in $U_R$ (in subdomain 2). As a consequence, for the conservative form, loss 2 is now represented by equation (\ref{eq:c_loss2}) with $MSE_{f_2}$ in equation (\ref{eq:C_vacuum_ur}). In this particular case, for both the conservative and the non-conservative forms, we pick $\delta_1=1$ and $\delta_2=0.8$. The result for rescaling cPINN in the conservative form is depicted in Figure \ref{fig:case4ab} by a cyan-dashed line, whereas the non-conservative form is shown by a pink-dashed line. The conservative and non-conservative forms' respective relative $L_2$ norms are $8.82 \cdot 10^{-2}$ and $5.99 \cdot 10^{-2}$. As we discovered based on the relative $L_2$ norm, cPINN's performance is not improved by applying the rescaling technique in the conservative form. Rescaling in the non-conservative form, however, only slightly improves the performance.

\subsubsection{Case 5a} \label{case5a}
\begin{figure}[ht]
    \centering\includegraphics[scale=0.45]{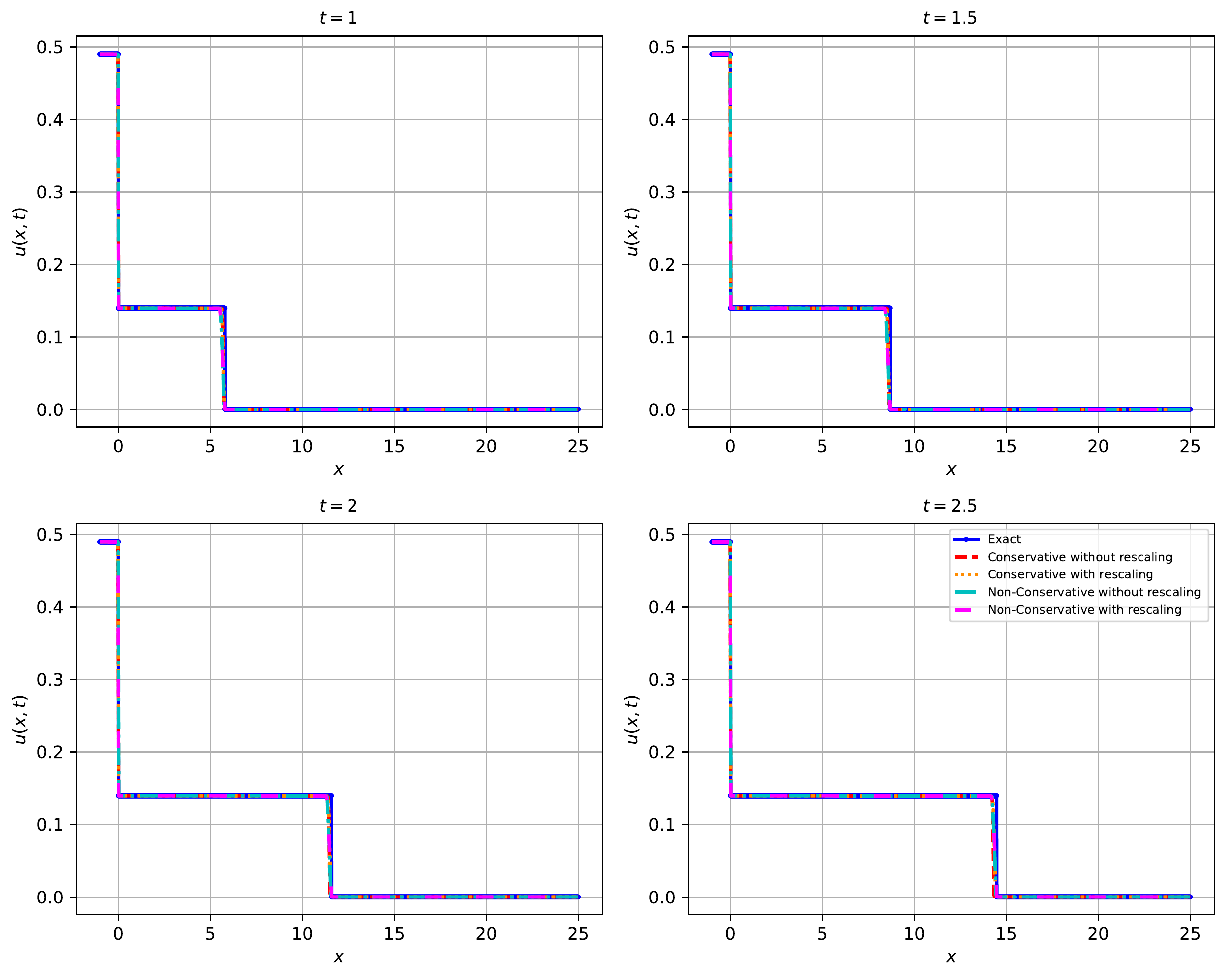}
    \caption{The evaluation of the exact and cPINN solutions is offered in both conservative and non-conservative forms, as well as without and with rescaling. For the conservative form, the relative $L_2$ norms are $7.87 \cdot 10^{-2}$ and $7.91 \cdot 10^{-2}$, respectively for without and with rescaling. On the other hand, the relative $L_2$ norms for the non-conservative form are $8.94 \cdot 10^{-2}$ and $6.57 \cdot 10^{-2}$, respectively, both without and with rescaling. As a result, only in its non-conservative form is rescaling able to improving cPINN performance.}
 \label{fig:case5ab}
\end{figure}

Another initial condition that we considered for the critical state occurs in $U_R$ is as follows:

\begin{equation} \label{eq:iv_case5_c}
    U_L = \begin{pmatrix}
    u_L \\ 
    \phi_L
    \end{pmatrix} = 
    \begin{pmatrix}
    0.49 \\ 
    0.7
    \end{pmatrix} \quad \text{ and } \quad
    U_R = \begin{pmatrix}
    u_R \\ 
    \phi_R
    \end{pmatrix} = 
    \begin{pmatrix}
    4 \cdot 10^{-4} \\ 
    0.2
    \end{pmatrix},
\end{equation}
for the conservative form. Additionally, the following is the initial condition for the non-conservative form when the critical state is in $\Tilde{U}_R$:
\begin{equation} \label{eq:iv_case5_nc}
    \Tilde{U}_L = \begin{pmatrix}
    \Tilde{u}_L \\ 
    \phi_L
    \end{pmatrix} = 
    \begin{pmatrix}
    0.7 \\ 
    0.7
    \end{pmatrix} \quad \text{ and } \quad
    \Tilde{U}_R = \begin{pmatrix}
    \Tilde{u}_R \\ 
    \phi_R
    \end{pmatrix} = 
    \begin{pmatrix}
    2 \cdot 10^{-3} \\ 
    0.2
    \end{pmatrix}.
\end{equation}
The domain for this case is $0 \leq t \leq 3$ and $-1 \leq x \leq 25$.
Moreover, the theoretical illustration is available in Figure \ref{fig:case6}. The outcomes of cPINN without the rescaling technique are shown in Figure \ref{fig:case5ab}; the red-dashed line and the cyan-dashed line, respectively, represent the results for the conservative and non-conservative forms. Both the conservative and non-conservative forms function sufficiently well in this situation.

\subsubsection{Case 5b} \label{case5b}
In this case, we use the rescaling technique to run cPINN in both the conservative and non-conservative forms using the identical domain and initial data from Section \ref{case5a}. In the conservative form, we choose the rescaling parameters as $\delta_1=1$ and $\delta_2=0.8$, whereas for the non-conservative form, $\delta_1=1$ and $\delta_2=0.4$ are selected. For the conservative and non-conservative forms, the relative $L_2$ norms are $7.91 \cdot 10^{-2}$ and $6.57 \cdot 10^{-2}$, respectively. As we observed, rescaling causes the performance worse in the conservative form while improving it in the non-conservative form. The results also represented in Figure \ref{fig:case5ab} as an orange-dots and pink-dashed line for the respective conservative and non-conservative forms.

The relative $L_2$ norm for each case, in both conservative and non-conservative forms, as well as for cases with and without the use of the rescaling approach, is contained in Table \ref{tab:relative_l2_norm}.

\begin{table}[ht]
\centering
\begin{tabular}{@{}p{0.25\linewidth}p{0.25\linewidth}p{0.25\linewidth}c@{}}
\toprule
\toprule
\multirow{2}{*}{} & \multirow{2}{*}{\textbf{Non-Critical}}                                                                           & \multicolumn{2}{c}{\textbf{Critical}}                                                                                                                                                                                                                                                                                  \\ \cmidrule(l){3-4} 
                  &                                                                                                         & \multicolumn{1}{c}{$U_L$}                                                                                                     & $U_R$                                                                                                                                                                         \\ \midrule \midrule
\textbf{Conservative}      & \begin{tabular}[c]{@{}c@{}}\textbf{Case 1:}\\ $8.96 \cdot 10^{-3}$\\ \textbf{Case 2:}\\ $1.11 \cdot 10^{-2}$\end{tabular} & \multicolumn{1}{c}{\begin{tabular}[c]{@{}c@{}} \textbf{Case 3a:}\\ $3.33 \cdot 10^{-1}$\\ \textbf{Case 3b:}\\ $1.73 \cdot 10^{-2}$\end{tabular}} & \begin{tabular}[c]{@{}c@{}} \textbf{Case 4a:}\\ $7.2 \cdot 10^{-2}$\\ \textbf{Case 4b:}\\ $8.82 \cdot 10^{-2}$\\ \textbf{Case 5a:}\\ $7.87 \cdot 10^{-2}$\\ \textbf{Case 5b:}\\ $7.91 \cdot 10^{-2}$\end{tabular}  \\ \midrule
\textbf{Non-Conservative}  & \begin{tabular}[c]{@{}c@{}} \textbf{Case 1:}\\ $6.05 \cdot 10^{-3}$\\ \textbf{Case 2:}\\ $8.82 \cdot 10^{-3}$\end{tabular} & \multicolumn{1}{c}{\begin{tabular}[c]{@{}c@{}} \textbf{Case 3a:}\\ $1.54 \cdot 10^{-2}$\\ \textbf{Case 3b:}\\ $2.21 \cdot 10^{-2}$\end{tabular}} & \begin{tabular}[c]{@{}c@{}} \textbf{Case 4a:}\\ $6.09 \cdot 10^{-2}$\\ \textbf{Case 4b:}\\ $5.99 \cdot 10^{-2}$\\ \textbf{Case 5a:}\\ $8.94 \cdot 10^{-2}$\\ \textbf{Case 5b:}\\ $6.57 \cdot 10^{-2}$\end{tabular} \\ \bottomrule
\end{tabular}
\caption{Relative $L_2$ norm from a number of cases, including both conservative and non-conservative forms as well as cases with and without rescaling.}
\label{tab:relative_l2_norm}
\end{table}

\subsection{Comparison with WENO5}\label{set:comp_weno5}
In this section, WENO5 and the performance of the conservative form of cPINN are examined. Component-wisely, we implement the system \eqref{sys:GBL} by coupling fifth-order WENO spatial discretization with third-order TVD Runge-Kutta time discretization \cite{weno1,weno2}. The relative $L_2$ norm for both approaches is displayed in Table \ref{table:c_cpinn_weno_all}. In conservative form, WENO5 outperforms cPINN, with the exception of cases where critical occurs in $U_R$ (Case 4 and 5), as shown by Table \ref{table:c_cpinn_weno_all}. WENO5 for the non-conservative form, however, is yet unresolved, whereas cPINN is very capable of addressing the issue.

\begin{table}[ht]
\centering
\begin{tabular}{p{0.3\linewidth}p{0.3\linewidth}p{0.3\linewidth}}
\toprule \toprule
        & WENO5                & cPINN                \\ \toprule \toprule
Case 1  & $1.85 \cdot 10^{-3}$ & $8.96 \cdot 10^{-3}$ \\ 
Case 2  & $2.9 \cdot 10^{-3}$  & $1.11 \cdot 10^{-2}$ \\ 
Case 3a & $6.93 \cdot 10^{-3}$ & $3.33 \cdot 10^{-1}$ \\ 
Case 3b & $6.93 \cdot 10^{-3}$ & $1.73 \cdot 10^{-2}$ \\ 
Case 4a & $1.61 \cdot 10^{-1}$ & $7.2 \cdot 10^{-2}$  \\ 
Case 4b & $2.53 \cdot 10^{-1}$ & $8.82 \cdot 10^{-2}$ \\ 
Case 5a & $2.25 \cdot 10^{-1}$ & $7.87 \cdot 10^{-2}$ \\ 
Case 5b & $3.16 \cdot 10^{-1}$ & $7.91 \cdot 10^{-2}$ \\ \hline
\end{tabular}
\caption{cPINN's relative $L_2$ norm in conservative form in comparison to WENO5's $L_2$ norm. As shown, WENO5 outperforms cPINN in cases 1-3 (both non-critical and critical situations in $U_L$), yet WENO5 is unable to handle the critical case that happens in $U_R$ (Case 4-5), whereas cPINN is able to handle this case.}
\label{table:c_cpinn_weno_all}
\end{table}

\section{Conclusions}
\label{sec:conclusions}

Both cPINN in the conservative and non-conservative forms of the generalized Buckley-Leverett perform admirably for the non-critical cases. Nevertheless, certain rescaling modifications must be made in order for cPINN to function properly for the critical state in $U_L$ in conservative form. Surprisingly, the non-conservative form can approach the solution satisfactorily when a critical condition occurs in the $U_L$ without the requirement for a rescaling procedure. Furthermore, rescaling doesn't actually improve the performance of cPINN in the conservative form for the critical case in $U_R$.  In contrast, the rescaling improves the performance of cPINN in the critical case in $U_R$ in a non-conservative form. Finally, we assess the performance of WENO5 and cPINN. When cPINN is employed in its conservative form, WENO5 outperforms it. Meanwhile, WENO5 in its non-conservative form remains an open issue, whereas cPINN can manage it exceedingly well. Therefore, based on our studies, it is apparent that cPINN can effectively handle generalized Buckley-Leveret in both conservative and non-conservative forms, as well as in non-critical and critical (with some rescaling modifications) states.

\appendix

\section{Existence of $u^{*}$}\label{set:u-star}
In this section, we solve the following equation which is the same as equation (\ref{eqn:u-star})
\begin{equation}\label{eqn:y-star}
    f_u(y,\phi_R) = \frac{f(y,\phi_R)-f(u_R,\phi_R)}{y-u_R}.
\end{equation}
Recall that $f(u,\phi) = \frac{u^2}{u^2+M(\phi-u)^2}$, and the denominator of $f$ is denoted by $D(u,\phi)$.
For $y\neq u_R$, equation (\ref{eqn:y-star}) can be simplified
\begin{equation}\label{eqn:simp-y}
    \frac{2y(\phi_R-y)}{D(y,\phi_R)} = \frac{\phi_R(y+u_R)-2u_Ry}{D(u_R,\phi_R)}.
\end{equation}
After a routine computation, equation (\ref{eqn:simp-y}) can be written as follows
\begin{equation}\label{eqn:3-poly-y}
    (\phi_R-2u_R)y^3 + (\phi_Ru_R+2u_R^2)y^2-(2\phi_Ru_R^2+\tilde{M}\phi_R^3)y+\tilde{M}\phi_R^3u_R=0,
\end{equation}
where $\tilde{M}=M/(M+1)$.
Due to the fact that $u_R$ is a zero of the cubic polynomial (\ref{eqn:3-poly-y}),
the cubic polynomial can be factorized.
Therefore we obtain
\begin{equation}
    q(y) = (y-u_R)\left[ (\phi_R-2u_R)y^2 + 2\phi_Ru_Ry - \tilde{M}\phi_R^3 \right].
\end{equation}
If $\phi_R=2u_R$, then $q(y)$ is reduced to a quadratic polynomial.
\begin{equation}
    q(y) = \phi_R^2(y-u_R)(y-\tilde{M}\phi_R).
\end{equation}
Clearly, the zero is $y=\frac{2Mu_R}{M+1}>u_R$ by $M>1$.

If $\phi_R\neq 2u_R$, $q(y)$ has two zeros other than $u_R$, denoted by $u_{\pm}$,
\begin{equation}
    u_{\pm} = \frac{1}{2(\phi_R-2u_R)}\left( -2\phi_Ru_R\pm\sqrt{ (2\phi_Ru_R)^2+4\tilde{M}(\phi_R-2u_R)\phi_R^3 } \right).
\end{equation}
Note that the discriminant is non-negative, since

\begin{eqnarray*}
     (2\phi_Ru_R)^2+4\tilde{M}(\phi_R-2u_R)\phi_R^3
     &=& 4\phi_R^2( u_R^2 -2\tilde{M}\phi_Ru_R+\tilde{M}\phi_R^2 ) \\
    &\geq& 4\phi_R^2( u_R^2 -2\tilde{M}\phi_Ru_R+\tilde{M}^2\phi_R^2 ) \\
    &=& 4\phi_R^2( u_R-\tilde{M}\phi_R)^2 \\ &\geq& 0.
\end{eqnarray*}
by the fact that $M>1$.
For the sub-case, $\phi_R>2u_R$.
We obtain that $q(y)$ has only one positive zero $u_{+}$ and it is small than $\phi_R$.
For the rest sub-case, $\phi_R<2u_R$.
$q(y)$ has two positive zeros, and $u_{+}<\phi_R<u_{-}$.
Hence as long as $\phi_R\neq 2u_R$, $u_{+}$ is the only one choice.
\bigskip

\section{Experiments Detail} \label{experiment_detail_appendix}

From the initial condition, we randomly selected 101 and 499 training points from SD1 and SD2, respectively. Moreover, using Latin Hypercube Sampling, 3000 points from SD1 are randomly selected for the interior. Depending on the cases, we sample 12500 or 17500 points for SD2. We use 12500 interior points for cases 1-3 and 17500 interior points for the remaining cases. Additionally, we arbitrarily selected 99 points from the interface, where the interface is always placed at $x=0.01$. We use the Glorot uniform distribution \cite{xavier} to initialize the parameters. Also, we implement tanh as the activation function in the hidden layer and sigmoid in the output layer. 

The following describes the MLP architecture. Whereas the number of neurons in each case is the same (40 neurons), the number of hidden layers varies by subdomain. In subdomain 1 we utilize eight hidden layers, while in subdomain 2 we use ten hidden layers.
 For the optimization technique, we deploy Adam \cite{adam}, and we train the neural network across 100,000 epochs. The initial learning rate is set to $10^{-3}$ and decreases linearly during training.
 \bigskip

\bibliographystyle{unsrt}
\bibliography{main.bib}

\end{document}